\documentclass{article}

\usepackage{arxiv}

\usepackage{multirow}
\usepackage{array}
\usepackage[ruled,vlined]{algorithm2e}

\usepackage[utf8]{inputenc} 
\usepackage[T1]{fontenc}    
\usepackage{hyperref} 
\usepackage{scalerel,stackengine}
\newenvironment{myProperty}{
   \raggedright
}

\stackMath
\newcommand\reallywidehat[1]{%
\savestack{\tmpbox}{\stretchto{%
  \scaleto{%
    \scalerel*[\widthof{\ensuremath{#1}}]{\kern-.6pt\bigwedge\kern-.6pt}%
    {\rule[-\textheight/2]{1ex}{\textheight}}
  }{\textheight}%
}{0.5ex}}%
\stackon[1pt]{#1}{\tmpbox}%
}

\usepackage{graphicx}
\usepackage[export]{adjustbox}
\usepackage{xcolor}
\usepackage{subfigure}
\usepackage{colortbl}
\usepackage{paralist}
\usepackage{bbm}
\usepackage{url}            
\usepackage{booktabs}       
\usepackage{amsfonts}       
\usepackage{nicefrac}       
\usepackage{microtype}      
\usepackage{lipsum}	
\usepackage{xcolor}
\title{Harnessing Code Switching to Transcend the Linguistic Barrier}

\author{
Ashiqur R. KhudaBukhsh\thanks{Ashiqur R. KhudaBukhsh and Shriphani Palakodety  are equal contribution first authors. Ashiqur R. KhudaBukhsh is the corresponding author.} \\
  Carnegie Mellon University \\
  \texttt{akhudabu@cs.cmu.edu} \\
  \And
  Shriphani Palakodety$^*$\\
  Onai\\
  \texttt{spalakod@onai.com} \\
\And
Jaime G. Carbonell \\
  Carnegie Mellon University\\
  \texttt{jgc@cs.cmu.edu} \\
}

\begin{document}
\maketitle

\begin{abstract}

Code mixing (or code switching) is a common phenomenon observed in social-media content generated by a linguistically diverse user-base. Studies show that in the Indian sub-continent, a substantial fraction of social media posts exhibit code switching. While the difficulties posed by code mixed documents to further downstream analyses are well-understood, lending visibility to code mixed documents under certain scenarios may have utility that has been previously overlooked. For instance, a document written in a mixture of multiple languages can be partially accessible to a wider audience; this could be particularly useful if a considerable fraction of the audience lacks fluency in one of the component languages. In this paper, we provide a systematic approach to sample code mixed documents leveraging a polyglot embedding based method that requires minimal supervision. In the context of the 2019 India-Pakistan conflict triggered by the Pulwama terror attack, we demonstrate an untapped potential of harnessing code mixing for human well-being: starting from an existing hostility diffusing \emph{hope speech} classifier solely trained on English documents, code mixed documents are utilized as a bridge to retrieve
\emph{hope speech} content written in a low-resource but widely used language - Romanized Hindi. Our proposed pipeline requires minimal supervision and holds promise in substantially reducing web moderation efforts.

\end{abstract}


\section{Introduction}

Analyzing geopolitical events through the lens of social media is a highly active research domain. From referendums with far-reaching political consequences (e.g., Brexit~\cite{celli2016predicting}) to sensitive and highly polarizing issues like mass shootings in the US~\cite{demszky-etal-2019-analyzing}, large scale social media analysis has the potential to offer important insights to political and social scientists. While social media discussions lend a great platform to exchange ideas, share opinions and debate issues, tackling online attacks targeted at certain individuals, or communities forms an important modern-day social media challenge to ensure human well-being.

Typical approach to moderate online hate consists of detecting \emph{hate speech} for subsequent moderation. However, one recent line of work argued in favor of identifying positive content in the context of heated online discussions between nuclear adversaries at the brink of waging full-fledged war~\cite{kashmir}. In a substantial corpus of 2.04 million YouTube comments on videos relevant to the 2019 India-Pakistan conflict triggered by the Pulwama terror attack,~\cite{kashmir} advocates the importance of hostility-diffusing comments and defines a new task of detecting hostility-diffusing \emph{hope speech}.    

While~\cite{kashmir} presents an important study of modern conflict between two nuclear adversaries with a long history of acrimonious past and grim projection of consequences should there be a full-blown war (100 million projected death as forecast by~\cite{toon2019rapidly}), typical to several studies conducted in linguistically diverse regions, the focus is largely restricted to the English subset of the comments (921,235 English comments posted by 392,460 users). With a combined language base of more than 500 million speakers of Hindi in India and Pakistan as opposed to nearly 250 million speakers in English, and a considerable fraction using Romanized Hindi on the web, extending this result to the Romanized Hindi subset of comments has understandable benefits. However, such omissions are common in social analyses due to lack of NLP (Natural Language Processing) resources in the native language. Most NLP  pipelines are monolingual - state of the art POS (Part-of-Speech) taggers, parsers, or NER (Named Entity Recognition)
taggers are rarely trained to handle multi-lingual documents and largely focus on English. Moreover, apart from typical colloquial style social media content, the presence of  code mixing\footnote{Throughout the paper, we use the terms code switching and code mixing interchangeably.}~\cite{gumperz1982discourse, myers1993dueling} -- seamless alteration between multiple languages within the same document boundary (e.g., a tweet, or a comment on a YouTube video) -- makes the task substantially more challenging. 




While the challenges posed by code switching to downstream analyses are well-documented, in this paper, we focus on a largely under-explored research question: \emph{How can code switching be harnessed for social good and human well-being?} 

Our research question is motivated by a simple intuition that a short text document is likely to express a consistent sentiment; if reliable linguistic separation of such code mixed documents can be achieved, the Hindi portion of the comments can be further harnessed to explore similar comments in the Hindi subset for which we require no further training (the \emph{hope speech} classifier we used in this paper is trained on English comments). Effectively, our method uses the Hindi portions of code mixed comments as a seed set of comments to mine similar content authored in Hindi. Our approach presents a compelling case study on how code switching can be harnessed as a bridge to detect peace-seeking content written in a low-resource language. A reliable system to identify code mixed \emph{hope speech} documents has additional untapped benefits. Intuitively, a code mixed document written in  two dominant languages in a linguistically diverse region is likely to be partially accessible to a wider set of audience.

\noindent\textbf{Contributions:} Our contributions are the following. 
\begin{compactenum}
\item\textbf{Human well-being:} We focus on an important task of detecting hostility-diffusing \emph{hope speech}~\cite{kashmir}. Social media would play an increasingly important role in understanding and analyzing modern conflicts~\cite{zeitzoff2017social}; online discussions between countries with a long history of conflicts is an under-studied yet highly important area. 

\item\textbf{Framework:} Code switching is typically viewed as an impediment to effective corpus analysis; to the best of our knowledge, we first highlight its untapped potential in acting as a bridge. While the role of mother tongue as a \emph{conversational lubricant} in a code switched environment has been previously studied in educational settings~\cite{butzkamm1998code}, harnessing code switching to effectively sample content from a sub-corpus written in a different language was never explored heretofore to our knowledge.      
\item\textbf{Machine Learning:} We leverage recent literature in language identification and Active Sampling to sample documents exhibiting high-levels of code mixing and provide an end-to-end pipeline to sample from Romanized Hindi starting with a \emph{hope speech} classifier trained on English documents. Our results indicate that our approach reduces considerable moderation efforts in detecting \emph{hope speech} written mostly in Romanized Hindi. 
\end{compactenum}

\section{Related Work}

Code switching has been a widely studied area in linguistics for nearly half a century~\cite{auer2013code}. While recent work on analyzing the social aspects of code mixing in online communities is gaining importance~\cite{yoder-etal-2017-code}, typically, code switching is viewed as an impediment to downstream NLP analyses and much of the focus in the community is concentrated in detecting token-level language and switch points~\cite{das2014identifying,rijhwani2017estimating,gella-etal-2014-ye} for cleaner linguistic separation. To the best of our knowledge, harnessing code switching for social good and human well-being has been largely unexplored. Our work draws inspiration from field-work in classroom settings showing how code switching helps students overcome linguistic barriers and how native tongue is used in a code mixed setting as a  ``conversational lubricant''~\cite{butzkamm1998code}.

Our work focuses on an important domain of online hostility-diffusion between civilians of nuclear adversaries first studied in~\cite{kashmir}. We use several resources presented in the paper (e.g., data set, language identification method with minor modification).  However, in~\cite{kashmir} the primary focus was mostly restricted to the English subset of comments, whereas in our work, we focus on leveraging an untapped potential of code switching and propose a pipeline to identify hostility-diffusing \emph{hope speech} from the Hindi sub-corpus, a task not addressed in~\cite{kashmir}. 

The importance of a robust token-level language identification system was explored in \cite{gella-etal-2014-ye}. The study demonstrated that typical document-level language
identification systems are a poor fit for code mixed documents. In the context of Indian
social media, \cite{gella-etal-2014-ye} also provided statistics on the use of Romanized
Hindi, and code-mixed text revealing significant use. Language preference priors for expression opinion were further investigated in ~\cite{rudra2016understanding} revealing
that negative opinion is often presented in Hindi, and~\cite{yoder-etal-2017-code} demonstrating
Wikipedia editor success.

Several studies have addressed challenges in analyzing code mixed text by using a token-level language-identification step in their NLP pipelines~\cite{23fcc7f84fd041ed827f51d17f745e4c,Elfardy2013SentenceLD,rijhwani-etal-2017-estimating}. \cite{rijhwani-etal-2017-estimating} in particular presents an HMM-based unsupervised token-level language-identification method to analyze code-switching statistics on social
media.

Recent studies have used sentence embeddings for sampling comments similar to
a ``query'' document. \cite{kumar2019submodular,Rohingya} used sentence embeddings in active learning settings to expand a training data set. In both cases
sampling based on the sentence embeddings yielded a far better rate of acquisition
of the desired class compared to random sampling. We utilize the polyglot embeddings
themselves as sentence embeddings in our rare positive mining task.

\section{Problem Definition}


\subsection{Task: \emph{Hope Speech} Detection}

We focus on the prediction task of \emph{hope speech} detection, first proposed in~\cite{kashmir} in the context of online discussions relevant to the 2019 India-Pakistan conflict. Aimed at diffusing hostility, a \emph{hope speech} classifier is a nuanced classifier (precise definition of \emph{hope speech} with illustrative examples is presented in~\cite{kashmir}) to detect content that contains a unifying message focusing on the war's futility, the importance of peace, and the human and economic costs involved, or expresses criticism of either the author's own nation's entities or policies, or the actions or entities of the two involved countries.

\noindent\textbf{Data set:} Our  data set, $\mathcal{D}$, consists of 2.04 million comments posted by 791,289 user on 2,890 YouTube videos relevant to this India-Pakistan conflict. Our main focus is on the English and Romanized Hindi subsets denoted as $\mathcal{D}_{\emph{en}}$ (921,235 comments) and $\mathcal{D}_{h_e}$ (1033,908 comments), respectively.    

\noindent\textbf{Annotated data set:} The \emph{hope speech} classifier is trained on an annotated data set, $\mathcal{D}_{\emph{hope}}^{\emph{train}}$, of 2,277 positive and 7,716 negative English comments and an in-the-wild performance (on data not belonging to the train or test set) of 84.68\% precision was reported.

\subsection{An Illustrative Example}

To motivate our intuition, we first provide an illustrative example of a code switched comment exhibiting \emph{hope speech} along with a loose translation. English, Hindi and neutral tokens (e.g., proper nouns, numerals, or technology terms) are color-coded with blue, red and black respectively (color scheme is consistent throughout the paper).

\begin{myProperty}
\noindent\rule{\textwidth}{1pt}
\small{\texttt{\textbf{\textcolor{blue}{I am} \textcolor{black}{Indian} \textcolor{blue}{and I say peace is the only solution} \textcolor{red}{ankh k badle ankh mangoge toh sari dunya andhi hojayegi}}}}
\noindent\rule{\textwidth}{1pt}
\small{\emph{I am Indian, and I say peace is the only solution; an eye for an eye makes the whole world blind.}}
\noindent\rule{\textwidth}{1pt}
\end{myProperty}

In the above example, both the Hindi and English components exhibit peace-seeking intent. Our main goal in this paper is to harness the Hindi components present in these highly code mixed \emph{hope speech} comments to detect \emph{hope speech} in the Hindi sub-corpus. Associated research questions are the following: 
\begin{compactitem}
    \item How can we sample code mixed documents? 
    \item How can we harness the Hindi part of a code mixed document to sample \emph{hope speech} from the Hindi portion? 
\end{compactitem}

\subsection{A Challenging Data Set}

Similar to most data sets of noisy, short social media texts generated in a linguistically diverse region, our data set exhibits a considerable presence of out-of-vocabulary (OOV) words, code mixing, and grammar and spelling disfluencies. In addition to these challenges, given that a vast majority of the content contributors do not speak English as their first language, we noticed varying levels of English proficiency in the corpus with a substantial incidence of phonetic spelling errors (e.g., [\textbf{\texttt{\textcolor{blue}{thankyou pakusta for hiumaniti no war} \textcolor{red}{aman ssnti kayam kare}}}] loosely translates to \emph{Thank you Pakistan for humanity; let peace prevail.}); 32\% of times, the word \texttt{liar} was misspelled as \texttt{lier}. Since Romanized Hindi does not have any standard spelling (e.g., the word \texttt{aman} meaning peace is spelled in the corpus as amun, amaan and aman), a high level of spelling variations added to the challenges. 


%

\noindent\textbf{How hard is it to sample \emph{hope speech}?} On a random sample of 1,000 comments from $\mathcal{D}_{h_e}$, our annotators\footnote{For all tasks, two annotators proficient in English, Hindi, Urdu, and Dutch were used. Across all rounds of labeling, the minimum Fleiss' $\kappa$ measure was high (0.84) indicating strong inter-rater agreement. After independent labeling, differences were resolved through discussion.} found 18 positives (i.e., 1.8\%). This result aligns with results reported in~\cite{kashmir} where only 2.45\% randomly sampled English comments were marked as \emph{hope speech}. Additionally, a previous study of a multilingual Hindi-English tweet corpus observed that Hindi was more commonly used to express negative sentiment~\cite{rudra2016understanding}. The minuscule presence of \emph{hope speech} indicates that  detecting such content is essentially a rare positive mining task and automated methods are essential. 

\subsection{Our Pipeline}

\noindent\textbf{Research question:} \emph{How to harness code switching to sample hope speech from the Hindi subset $\mathcal{D}_{h_e}$?}

\begin{figure}[htb]
\centering
\includegraphics[scale = 0.3]{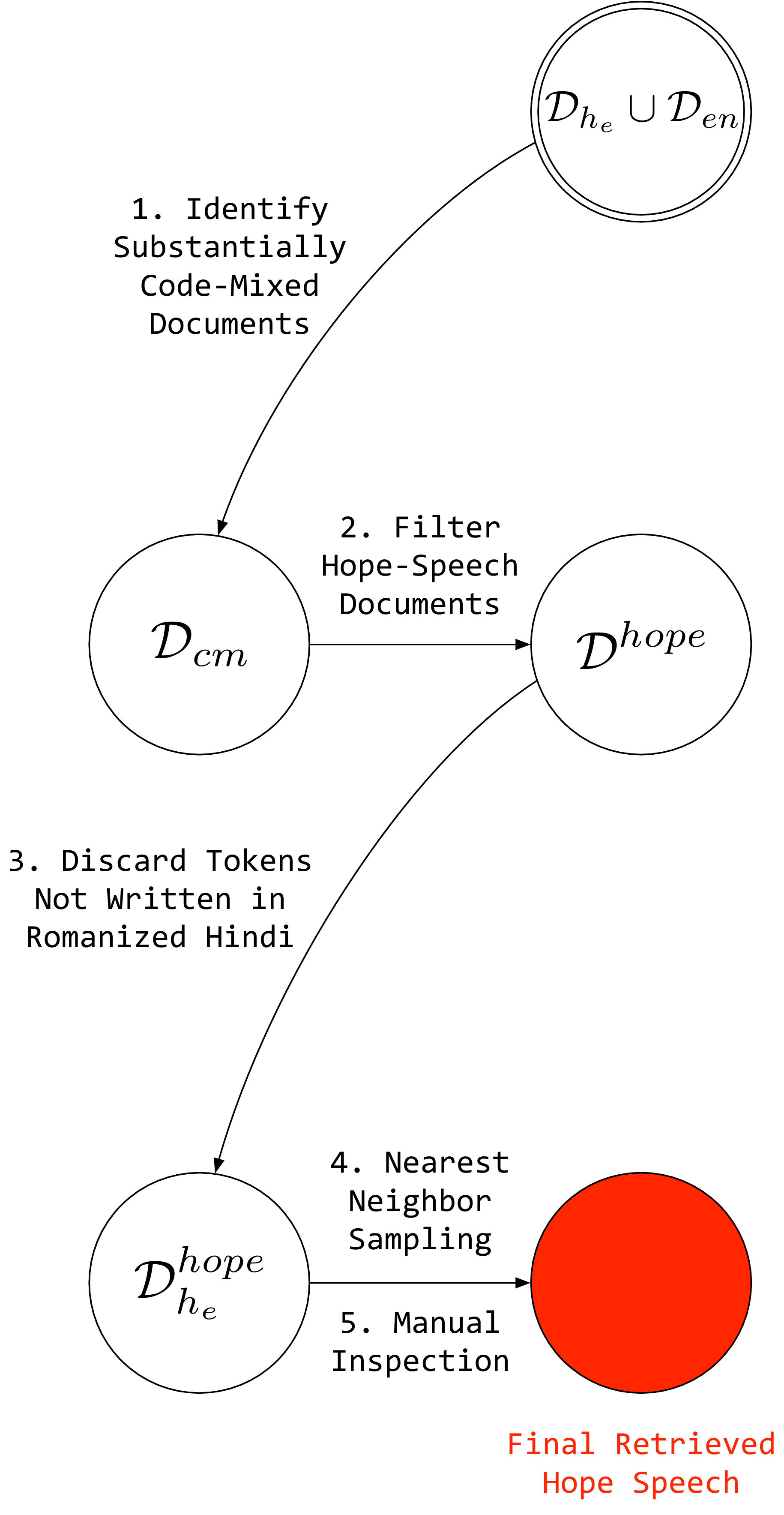}
\caption{System diagram.}
\label{fig:ALSystemFlow}
\end{figure}

A schematic diagram of our pipeline to sample \emph{hope speech} from the Hindi subset, $\mathcal{D}_{h_e}$, is presented in Figure~\ref{fig:ALSystemFlow}. 
Our pipeline consists of the following steps.

\begin{compactenum}
\item Identify the subset, $\mathcal{D}_{cm}$, from $\mathcal{D}_{h_e} \cup \mathcal{D}_{\emph{en}}$ with substantial code mixing.
\item Run the \emph{hope speech} classifier (trained on annotated English comments $\mathcal{D}_{\emph{hope}}^{\emph{train}}$) on $\mathcal{D}_{\emph{cm}}$ and construct the subset $\mathcal{D}^{\emph{hope}}$ containing comments predicted as \emph{hope-speech}.
\item Construct $\mathcal{D}^{\emph{hope}}_{h_e}$ transforming each comment in $\mathcal{D}^{\emph{hope}}$ discarding any tokens not written in Romanized Hindi. 
\item Using $\mathcal{D}^{\emph{hope}}_{h_e}$ as the seed set, retrieve
the nearest neighbors in the comment embedding space from $\mathcal{D}_{h_e}$.
\item Manually inspect the obtained sampled comments to detect \emph{hope speech}. 

\end{compactenum}

Steps 1, 2, 3, and 4 require minimal manual supervision. Step 5 is the only step that requires substantial manual effort. Our results indicate that we obtained a nearly 10-fold improvement over our baseline.

\section{Methods and Results}

\noindent\textbf{Research question:} \emph{how to sample code mixed documents?}


\subsection{Code Mixing Index (CMI)}

We used a well-known metric to measure the extent of code switching in a document - Code Mixing Index (\emph{CMI}) - first proposed in~\cite{das2014identifying}. Essentially, \emph{CMI} measures the presence of a dominant language in a document. Let a document $d$ expressed with $k$ different languages, $\{l_1,\ldots,l_k\}$, and $u$ neutral tokens be represented as a sequence of words: $[w_1,\ldots,w_n]$. Let $\mathcal{L}$($w_i$) return the language of word $w_i$ (or neutral if it is a neutral token). For each language, $\mathcal{N}$($l_j$) denotes the total number of utterances of $l_j$ in the document, i.e., $\mathcal{N}$($l_j$) = $\sum_{i=1}^{i=n} \mathbbm{I}(\mathcal{L}(w_i) = l_j$) where $\mathbbm{I}$ is the indicator function. The \emph{CMI} of the document $d$, \emph{CMI}($d$), is measured as:\\ \emph{CMI}($d$) = $\frac{\sum_{j=1}^{j=k} \mathcal{N}(l_j) - max_i{(\mathcal{N}(l_i))}}{n - u}$. In the boundary condition, where every word in the document is a neutral token, \emph{CMI} is defined as 0; hence, \emph{CMI}($d$) $\in [0,1]$. A low \emph{CMI} value indicates minimal code switching i.e. the document is almost entirely written in the dominant language. Understandably, when $k = 2$, the highest possible \emph{CMI} is 0.5 indicating equal presence of two component languages. When  $\mathcal{L}$(.) is estimated using a language identification method, we denote the estimated \emph{CMI} of a document as $\reallywidehat{\emph{CMI}}(d)$. 

We now illustrate with an example: \texttt{[\textbf{\textcolor{red}{bilkul sahi baat kahi aapne} imran khan \textcolor{red}{saab} \textcolor{blue}{please please no more war only peace}}]} (loosely translates to \emph{You've spoken the absolute truth Mr. Imran Khan, please no more war, only peace.}). In this example, $\mathcal{N}(en) = 7$, $\mathcal{N}(h_e) = 6$, $n$ = 15, and $u$ = 2. Hence, \emph{CMI} of the document is $\frac{7 + 6 - 7}{15 - 2}$ = 0.46. We considered documents with $\reallywidehat{\emph{CMI}}$ greater than or equal to 0.4 as documents exhibiting significant code mixing.

\begin{figure}[t]
\vspace{-0.38cm}
\centering
\includegraphics[trim={0 0 0 0},clip, height=1.2in]{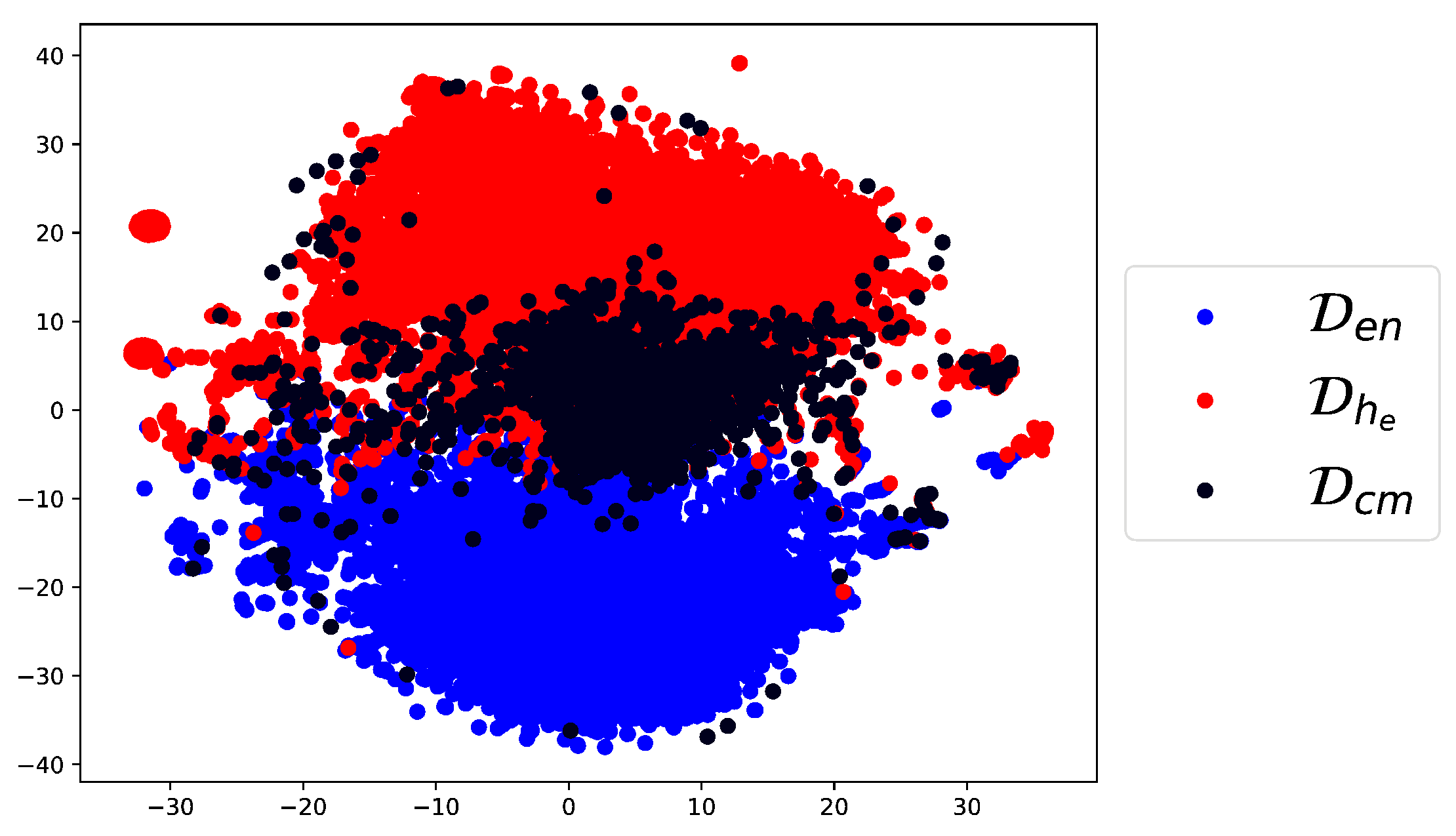}
\caption{\small{A TSNE ~\cite{maaten2008visualizing} plot of the polyglot document-embedding space. The code-mixed region (black) lies between the Hindi (red) and English (blue) language clusters.}}
\label{fig:cluster}
\end{figure}

\subsection{Estimating CMI}

In order to sample documents with high \reallywidehat{\emph{CMI}}, we need a reliable token-level language identification module. We used the polyglot-embedding based method proposed in~\cite{kashmir} (we denote this method by $\hat{\mathcal{L}}_{\emph{polyglot}}$). We chose $\hat{\mathcal{L}}_{\emph{polyglot}}$ because it requires minimal supervision and is particularly well-suited for noisy social media texts~\cite{Rohingya}. In particular, $\hat{\mathcal{L}}_{\emph{polyglot}}$ involves obtaining the document embeddings, and then using $k-$Means on these embeddings. The method is shown to reveal highly precise language clusters. Previous use-cases of $\hat{\mathcal{L}}_{\emph{polyglot}}$ were limited to document-level language identification. In our experiments we found that without any significant modification, the technique is capable of token-level language identification with considerable accuracy. Our token-level language identification follows the same method presented in~\cite{kashmir}. We consider a token as a single-word document, obtain its embedding, and assign language to the nearest cluster center in the document embedding space. 

\begin{table}[t]
\small
\centering
\begin{tabular}{|p{10cm}|}
\hline
 Pakistan, 
 he, 
 army,
 media, 
 Modi, 
 Pak, 
 Pakistani, 
 Kashmir, 
 pilot, 
 attack, 
 video, 
 news, 
 khan,
 jai, 
 2,
 hind,
 Imran,
 Muslim,
 sir,
 1
 \\ \hline
\end{tabular}
\caption{\small{Top 20 neutral words by frequency detected by $\hat{\mathcal{L}}_{\emph{polyglot}}$.}}
\label{tab:neutral}
\end{table}

\noindent\textbf{Detecting neutral tokens:} Neutral tokens are identified using a simple heuristic: for a two-language scenario, a token is marked neutral if it is approximately equidistant from the two respective cluster centers. For a given token, $w$, let the Euclidean distance of $w$ from the English cluster and Hindi cluster in the comment embedding space be represented as $\emph{dist}(w,en)$ and $\emph{dist}(w,h_e)$, respectively. Let the distance between the two cluster centers be expressed as $\emph{dist}(en,h_e)$.  $\hat{\mathcal{L}}_{\emph{polyglot}}(w)$ = $\emph{neutral}$ iff 
$w \in (\mathcal{D}_{h_e} \cup \mathcal{D}_{\emph{en}})$ and $\frac{|\emph{dist}(w,en) - \emph{dist}(w,h_e)|}{\emph{dist}(en,h_e)} \le \epsilon$.

Table~\ref{tab:neutral} shows the top 20 (ranked by frequency) neutral tokens detected by our method when $\epsilon$ is set to 0.1. They broadly include proper nouns (e.g., Modi, Khan, Pakistan), numerals (e.g., 1), technical terms (e.g., video) and overloaded words (e.g., he; he in Hindi is the verb \emph{is}, and is the third-person singular masculine pronoun in English).

\begin{table*}[htb]
{
\scriptsize
\begin{center}
     \begin{tabular}{|p{0.40\textwidth}|p{0.40\textwidth}|}
     \hline
     Code switched \emph{hope speech} & Loose translation \\
    \hline
\textbf{\texttt{\textcolor{blue}{I am} \textcolor{black}{Pakistani} \textcolor{red}{agar ap dono} \textcolor{blue}{country} \textcolor{red}{ny} \textcolor{blue}{war} \textcolor{red}{karne ha to gurbat khatam karnay ke} \textcolor{blue}{war} \textcolor{red}{karo dono} \textcolor{blue}{countries} \textcolor{red}{bht gareeb hain} \textcolor{blue}{plzz dont do war war is not solution of peace}}} &     \emph{I am Pakistani. If both countries have to wage a war, wage a war to end poverty; both countries are very poor. Please, do not war, war is not solution of peace.}\\
\hline
\textbf{\texttt{\textcolor{blue}{please media} \textcolor{red}{walo nafrat phailana chhod do} \textcolor{blue}{we want peace only} \textcolor{red}{jai hind}}} & \emph{Please media folks, stop spreading hate. We want peace; hail India!}\\
\hline
\texttt{\textbf{\textcolor{blue}{absolutely right i think} \textcolor{red}{ab netao kee jung ko ham dono mumalik ne rad karna hai} \textcolor{blue}{we people of both countries want peace peace and peace}}} & \emph{Absolutely right. I think politicians' war has to prevented by common people of both countries. We people of both countries want peace, peace, and peace.}\\
    \hline
    \end{tabular}
    
\end{center}
\caption{{Random sample of code mixed \emph{hope speech} obtained by \emph{hope speech} classifier run on $\mathcal{D}_{\emph{cm}}$.}}
\label{tab:classifierPerf}}
\end{table*}

\begin{table*}[htb]
{
\scriptsize
\begin{center}
     \begin{tabular}{|p{0.40\textwidth}|p{0.40\textwidth}|}
     \hline
     Code switched \emph{hope speech} & Loose translation \\
    \hline
\textbf{\texttt{    
\textcolor{red}{bhai ap bhi khuch rahiye} allah \textcolor{red}{apki har farmaish puri kare} \textcolor{blue}{and I repeat again} \textcolor{red}{bhai} \textcolor{blue}{I love you all my dear brothers and sisters} \textcolor{red}{mujhe ekh dusre se} indian \textcolor{blue}{or} pakistaani \textcolor{red}{keh kar bulana bilkul pasand nahi hum sab bhai or bhen hai or rahenge} \textcolor{blue}{we are good humans of earth}}} 
 &     \emph{Brother, you also be happy. May Allah grant all your wish and I repeat again, brother, I love you all my dear brothers and sisters. I don't like to identify each other as Indian or Pakistani, we are all brothers and sisters, we are good humans of earth.} \\
\hline
 \textbf{\texttt{\textcolor{red}{galiyan dene se kya hoga beach me to begunah awam mare gii orr ham jo jang jang krte henn kha jang itnii asan he} \textcolor{blue}{no this war is end of the world because} Ind \textcolor{blue}{an} Pak \textcolor{blue}{is newclear states}}} & \emph{Nothing will come out of abusing, we will cry for war while innocent civilians will die. Who said that war is easy? No, this war is end of the world because Ind and Pak are nuclear states.}\\
\hline
\textbf{\texttt{\textcolor{red}{daikhou dosto apaas me baahss bazii maat kro} \textcolor{blue}{plz} \textcolor{red}{such me} \textcolor{blue}{I have love for both} Pakistan \textcolor{blue}{and} India \textcolor{red}{bus apaas me muhabaatey rakhou} \textcolor{blue}{I m student of} 9th \textcolor{red}{lakn me muhabaat chataa hu donoo} \textcolor{blue}{countries} \textcolor{red}{me choroo fazoul ki nafrateey} \textcolor{blue}{love you} Pakistan \textcolor{blue}{and my neighbour country} India}} & \emph{Look friends, stop quarrelling among each other. Please, for real, I have love for both Pakistan and India, just harbor love between each other. I am a student of 9th grade, but I want love between both countries. Leave this useless hate, love you Pakistan and my neighbor country India.}\\
    \hline
    \end{tabular}
    
\end{center}
\caption{{Random sample of \emph{hope speech} obtained through \emph{NN-sample}($\mathcal{D}^{\emph{hope}}$).}}
\label{tab:codeMixedHindiNotKilled}}
\end{table*}  

\begin{table}[htb]
\centering
\small
\begin{tabular}{|c|c|c|c|}
\hline  
    Corpus & \emph{CMI} & \reallywidehat{\emph{CMI}} & Overall RMSE \\
\hline  
    $\mathcal{D}_{\emph{en}}$   & 0.03 & 0.04 & \multirow{3}{1cm}{0.05} \\
\cline{1-3} 
    $\mathcal{D}_{{h_e}}$     & 0.10 & 0.12 & \\
\cline{1-3} 
   $\mathcal{D}_{\emph{cm}}$    & 0.38 & 0.45 & \\
\hline
\end{tabular}
\label{tab:CMIEstimation}
\caption{\small{\emph{CMI} estimation root mean squared error.}}

\end{table}

\begin{table*}[htb]
{
\scriptsize
\begin{center}
     \begin{tabular}{|p{0.40\textwidth}|p{0.40\textwidth}|}
     \hline
     Sampled \emph{hope speech} & Loose translation \\
    \hline
\textbf{\texttt{\textcolor{red}{khuda kare dono mulko ke beech aman ho jaye hum yahi chahate hai}}}
 &     \emph{God willing, peace between two countries happens. I want that}.\\
\hline
\textbf{\texttt{Europe \textcolor{red}{main sub aik sath rahe aur kitne mulk aik sath rehte hain aur aik hum hai ka larte hi rahe ga kab samjhe ga hum}}} & \emph{In Europe, so many countries are living together. Whereas we are still fighting, when will we understand?} \\
\hline
\textbf{\texttt{\textcolor{red}{koi mulk nhi chahta aur na kisi mulk ki awaam chaahti ki uske padosi mulk se ki aapsi taaluqaat kharaab ho}}} & \emph{No country or its civilians want that the relationship with its neighboring country sours.}\\
    \hline
    \end{tabular}
    
\end{center}
\caption{{Random sample of \emph{hope speech} obtained through \emph{NN-Sample}($\mathcal{D}^{\emph{hope}}_{h_e}$).}}
\label{tab:codeMixedHindiKilled}}
\end{table*}

\begin{table}[htb]
\centering
\scriptsize
\setlength{\extrarowheight}{2pt}
\begin{tabular}{cc|c|c|c|}
  & \multicolumn{1}{c}{} & \multicolumn{3}{c}{Predicted Label} \\
  & \multicolumn{1}{c}{} & \multicolumn{1}{c}{$neutral$}  & \multicolumn{1}{c}{$en$}  & \multicolumn{1}{c}{$h_e$} \\\cline{3-5}
            & $neutral$ &\cellcolor{blue!25} 702 & 325 & 144 \\ \cline{3-5}
True Label  & $en$ & 334 &\cellcolor{blue!25} 4690 & 56 \\\cline{3-5}
            & $h_e$ & 85 & 148 &\cellcolor{blue!25} 3235 \\\cline{3-5}
\end{tabular}
\caption{Confusion matrix of token-level performance evaluation of $\hat{\mathcal{L}}_{\emph{polyglot}}$ on 300 annotated comments from $\mathcal{D}_{h_e} \cup \mathcal{D}_{\emph{en}}$.}
\label{tab:performance}
\end{table}
 
On a data set of 300 comments with gold standard token-level annotation, we found that $\hat{\mathcal{L}}_{\emph{polyglot}}$ performs token-level language detection with considerable accuracy. As shown in Table~\ref{tab:performance}, the overall accuracy of $\hat{\mathcal{L}}_{\emph{polyglot}}$ is 88.76\%. 


\noindent\textbf{Estimating CMI:} Once the reliability of token-level language identification by $\hat{\mathcal{L}}_{\emph{polyglot}}$ is established, we next evaluate the reliability of the estimate for $\emph{CMI}$ (i.e. \reallywidehat{\emph{CMI}}). We first define the subset with substantial estimated code mixing, $\mathcal{D}_{cm}$, as the following:   $\mathcal{D}_{\emph{cm}}$ = $\{d\}$ s.t.    $d \in \mathcal{D}_{h_e} \cup \mathcal{D}_{en}$ and \reallywidehat{{}\emph{CMI}}($d$) $\ge 0.4$, i.e., the document is either part of the English or Romanized Hindi subset and its estimated \emph{CMI} is high indicating nearly equal presence of Hindi and English. As shown in Figure~\ref{fig:cluster}, of $\mathcal{D}_{\emph{cm}}$ indeed falls in the overlapping region of the English and Hindi cluster. We manually inspected and annotated 1,000 randomly sampled comments from $\mathcal{D}_{cm}$ and found that 95.9\% comments exhibited code switching. We further obtained token level consensus labels for 100 randomly sampled comments each from $\mathcal{D}_{cm}$, $\mathcal{D}_{en}$, and $\mathcal{D}_{h_e}$. Table~\ref{tab:CMIEstimation}  compares the ground truth \emph{CMI} and \reallywidehat{\emph{CMI}} and demonstrates that we achieved a reasonable approximation of true \emph{CMI} using $\hat{\mathcal{L}}_{\emph{polyglot}}$.

\subsection{Sampling \emph{Hope Speech} From $\mathcal{D}_{h_e}$}

\noindent{\textbf{Research question:}} \emph{How to harness the Hindi part of a code mixed document to sample \emph{hope speech} from  $\mathcal{D}_{h_e}$?}

Once we identify a comment subset with substantial code mixing, $\mathcal{D}_{\emph{cm}}$, obtaining \emph{hope speech} comments using an off-the-shelf \emph{hope speech} classifier is straight-forward. Out of 36,969 comments in $\mathcal{D}_{\emph{cm}}$, the classifier predicted a set of 199 comments, $\mathcal{D}^{\emph{hope}}$, as positives. Upon manual annotation, we obtained 149 positives (denoted as $\mathcal{D}^{\emph{hope}}_{+}$), i.e., 74.87\% positives. Understandably, due to presence of code switching, the in-the-wild precision in $\mathcal{D}_{\emph{cm}}$ is lower than previously reported~\cite{kashmir} in-the-wild precision of 84.68\% in $\mathcal{D}$. Table~\ref{tab:classifierPerf} lists a subset of randomly sampled comments from $\mathcal{D}^{\emph{hope}}_{+}$. We noted that the Hindi component of the comments were consistent with the overall sentiment of the comment. 

A noisy approximation of the Hindi sub-part of these comments can be obtained by $\hat{\mathcal{L}}_{\emph{polyglot}}$ through discarding non-Hindi tokens. 

\begin{myProperty}
\noindent\rule{\textwidth}{1pt}
\small{\texttt{\textbf{\textcolor{blue}{I love} India \textcolor{blue}{I am} Pakistani \textcolor{red}{mein amun chahta hon khuda ke waste jang nai} \textcolor{blue}{peace peace peace}}}}
\noindent\rule{\textwidth}{1pt}
\small{\emph{I love India, I am Pakistani. I want peace for God's sake, not war, peace peace peace.}}
\noindent\rule{\textwidth}{1pt}
\end{myProperty}

For instance, the above comment is transformed into [\textbf{\texttt{\textcolor{red}{mein amun chahta hon khuda ke jang nai}}}] (loosely translates to \emph{I want peace for God not war}) when we discard non-Hindi tokens using $\hat{\mathcal{L}}_{\emph{polyglot}}$. \texttt{Waste} is both a valid English and Hindi word (meaning \emph{sake}), and the language detector makes an error in correctly predicting it. We admit that it is possible to use more sophisticated methods to extract Hindi that consider context (e.g., considering context to assign label to a fence word) and possibly squeeze more performance out of it. However, we are primarily interested in establishing a blue-print for harnessing code switching for social good and testing the robustness of our pipeline without  resorting to performance-driven engineering. In every step of our pipeline, a better-performing algorithm (e.g., better language detection module, sophisticated method to extract Hindi, more powerful comment embeddings, further effective sampling technique) can be plugged in without disturbing the flow and with a possibility of performance improvement.

\noindent\textbf{Active Sampling:} Once we extract the Hindi sub-parts of $\mathcal{D}^{\emph{hope}}$ (denoted as $\mathcal{D}^{\emph{hope}}_{h_e}$), our next task is to find comments in $\mathcal{D}_{h_e}$ that are similar to the Hindi sub-part. To this end, we use a recently-proposed~\cite{Rohingya} Active Sampling algorithm which samples nearest neighbors in the comment embedding space to identify rare positives. Our choice of this Active Sampling technique is motivated by its effectiveness in mining rare positives and reported robustness to spelling variations which is particularly critical because our corpus contains noisy social media texts and Romanized Hindi does not have standard spelling rules.  Following~\cite{Rohingya}, we used cosine distance of the embeddings as the distance measure. 

Our sampling algorithm is described in Algorithm~\ref{algo:NN-sampling}. This algorithm takes a seed set, $\mathcal{S}$, and a sample pool $\mathcal{U}$ as inputs and outputs a set, $\mathcal{E}$ $\subset$ $\mathcal{U}$, containing nearest neighbors of $\mathcal{S}$ in the comment-embedding space.  Initially, our expanded set, $\mathcal{E}$, is an empty set. At each step, we expand this set with nearest neighbors that are already not present in the expanded set or the seed set. The function \emph{getNearestNeighbor}(\emph{c, dist}) returns the comment in $\mathcal{U}$ with minimum distance greater than or equal to \emph{dist}. The $\emph{size}$ parameter is set to 5, i.e., for each comment, we add five unique nearest neighbors. We set $\mathcal{U}$ to $\mathcal{D}_{h_e}$ since we are interested in detecting \emph{hope speech} in Hindi. 

\noindent\textbf{Baselines:}  Recall that, a random sample of 1,000 comments from $\mathcal{D}_{h_e}$ only yielded 1.8\% positives which is our primary baseline method (denoted as \emph{random-Sample}($\mathcal{D}_{h_e}$)). 

\begin{algorithm}[t]
\scriptsize
\DontPrintSemicolon
\SetAlgoLined

\textbf{Initialization:}\; 
$\mathcal{E} \leftarrow \{\}$\;

\textbf{Main loop:}\;
\ForEach{\emph{comment} ~c~$\in$~$\mathcal{S}$ }
{
$count \leftarrow 0$\;
$dist \leftarrow 0$\;
\While{count $\leq$ size}
{
    $neighbor \gets getNearestNeighbor(c, dist)$\;
    $dist \gets cosineDistance(c, neighbor)$\;
    
    \If{neighbor $\notin$ $\mathcal{E} \cup \mathcal{S}$}{$\mathcal{E} \gets \mathcal{E} \cup \{neighbor\}$\;
    $count \gets count + 1$\;}
  }
}
\textbf{Output:} $\mathcal{E}$
\caption{{ \small{\texttt{NN-Sample}($\mathcal{S}$, $\mathcal{U}$)}}}

\label{algo:NN-sampling}

\end{algorithm}

\begin{table}[htb]
\small
{

\begin{center}
     \begin{tabular}{|l | c  |}
     \hline 
     Method & Performance  \\
    \hline
     
  \emph{random-Sample}($\mathcal{D}_{h_e}$) & 1.8\%  \\
   \hline
   \emph{NN-Sample}($\mathcal{D}^{\emph{hope}}_{h_e}$) & 18.59\%  \\
   \hline 
  \emph{NN-Sample}($\mathcal{D}^{\emph{hope}}$) & 26.93\%  \\
   \hline 
   \emph{NN-Sample}($\mathcal{D}^{\emph{hope}}_{h_e, +}$) & 21.88\%  \\
   \hline 
  \emph{NN-Sample}($\mathcal{D}^{\emph{hope}}_{+}$) & \textbf{31.68\%}  \\
   \hline

    \end{tabular}

\end{center}
\caption{\small{Sampling performance.}}
\label{tab:samplingPerformance}}
\end{table}

\begin{table}[htb]
\scriptsize
\begin{center}
     \begin{tabular}{|l | c  |}
     \hline 
     Method & Performance  \\
    \hline
     
 $\mathcal{D}_{h_e}$ & 0.12  \\
   \hline
   $\mathcal{D}_{\emph{en}}$ & 0.04  \\
   \hline
   $\mathcal{D}_{\emph{hope}}^{\emph{train}}$ & 0.03  \\
   \hline 
  $\mathcal{D}_{\emph{cm}}$ & 0.44  \\
   \hline 
   \emph{NN-Sample}($\mathcal{D}_{\emph{hope}}^{h_e}$) & 0.05  \\
   \hline 
  \emph{NN-Sample}($\mathcal{D}_{\emph{hope}}$) & 0.43  \\
   \hline

    \end{tabular}

\end{center}
\label{tab:CMIOfRegions}
\caption{\small{Estimated CMI  comparison. }}
\end{table}

\begin{figure}[htb]
\vspace{-0.38cm}
\centering
\includegraphics[trim={0 0 0 0},clip, height=1.2in]{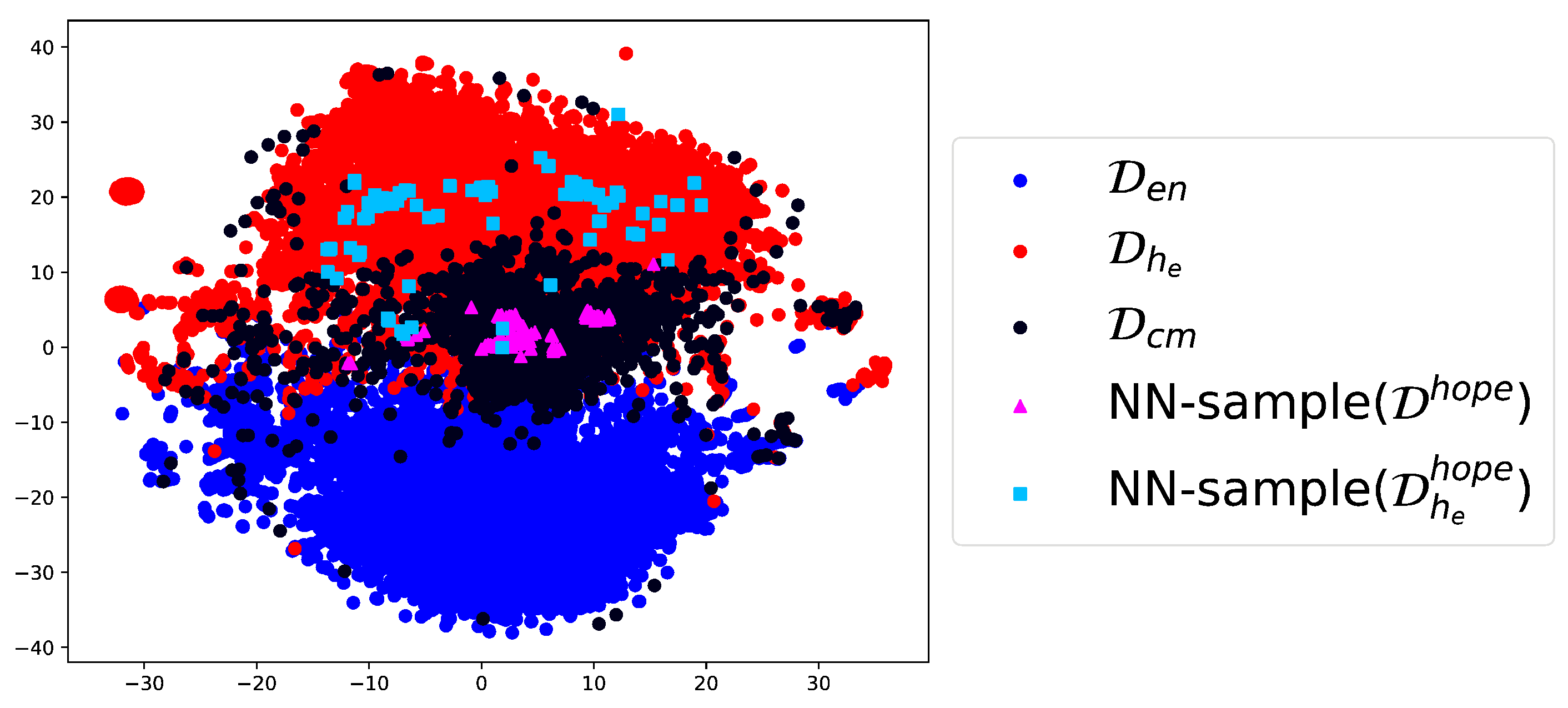}
\caption{\small{A 2D visualization showing the sampling results against the embedding space. Discarding non-Hindi tokens retrieves documents with low \emph{CMI} written mostly in Hindi.}}
\label{fig:cluster2}
\end{figure}

Table~\ref{tab:samplingPerformance} compares the performance of our sampling method against the baseline (we do not explicitly mention $\mathcal{U}$ which is consistently set to $\mathcal{D}_{h_e}$ across all \emph{NN-Sample} methods). We obtained substantial improvement over the baseline. Both \emph{NN-Sample}($\mathcal{D}^{\emph{hope}}_{h_e}$) and \emph{NN-Sample}($\mathcal{D}^{\emph{hope}}$) require human inspection only at the last step of our pipeline. Our results indicate that our approach can substantially reduce manual moderation effort in detecting \emph{hope speech}. Effectively, we sampled \emph{hope speech} from a Hindi corpus simply relying on a classifier trained on English comments and harnessing code switching as a bridge. In all steps of the pipeline, we perform noisy approximations in estimating $\emph{CMI}$, extracting Hindi sub-parts of comments and of course, detecting $\emph{hope speech}$. If we introduce little more supervision and instead expand the manually annotated \emph{hope speech} set $\mathcal{D}^{\emph{hope}}_+$, as expected, our performance improved. Our results indicate that using minimal manual supervision we can sample with as high as 30\% accuracy from the Hindi subset $\mathcal{D}_{h_e}$. 

\noindent\textbf{Research question:} \emph{What is the benefit of extracting the Hindi sub-part?} Both \emph{NN-Sample}($\mathcal{D}^{\emph{hope}}_{h_e}$) and \emph{NN-Sample}($\mathcal{D}^{\emph{hope}}_{h_e, +}$) are outperformed by 
corresponding sampling methods \emph{NN-Sample}($\mathcal{D}^{\emph{hope}}$) and \emph{NN-Sample}($\mathcal{D}^{\emph{hope}}_{+}$), respectively (see, Table~\ref{tab:samplingPerformance}). We were curious to analyze if  extracting the Hindi allows sampling from the sub-region of $\mathcal{D}_{h_e}$ 
mostly written in pure Hindi. As shown in Table~\ref{tab:CMIOfRegions}, without removing the non-Hindi part, \emph{NN-Sample} (intuitively) yielded a set with high level of code mixing. 
Nearest neighbors of a code switched comment are likely other code switched comments. However, sampling using just the Hindi sub-part yielded substantially less code mixing - Table ~\ref{tab:codeMixedHindiKilled} shows considerably less code-mixing than Table~\ref{tab:codeMixedHindiNotKilled}. Our intuition that removing non-Hindi is crucial for sampling from the low \emph{CMI} region of $\mathcal{D}_{h_e}$ is further supported by Figure~\ref{fig:cluster2} wherein  \emph{NN-Sample}($\mathcal{D}^{\emph{hope}}_{h_e}$) has a better spread over $\mathcal{D}_{h_e}$ while $\mathcal{D}^{\emph{hope}}$ is mostly located in the code mixed region.


\section{Conclusion}

In NLP literature, typically, code switching is viewed as an impediment to downstream analyses. In this paper, we first raise a novel proposition that code switching can be harnessed for social good and human well-being by using it as a bridge to retrieve hostility-diffusing content written in a low-resource language. Our approach is appealing for its minimal supervision requirements. In the context of hostility diffusing \emph{hope speech} comments, our methods can be used to broaden the reach
of such content overcoming the varied language skills of linguistically diverse regions and transcending language barriers.

\bibliographystyle{unsrt}

\newpage

\end{document}